\pdfoutput=1
\documentclass[11pt]{article}
\usepackage[final]{acl}

\usepackage{times}
\usepackage{latexsym}

\usepackage[T1]{fontenc}
\usepackage[utf8]{inputenc}
\usepackage{microtype}
\usepackage{inconsolata}
\usepackage{graphicx}
\usepackage{enumitem}
\usepackage{listings}
\usepackage{xcolor}
\usepackage{booktabs}
\usepackage{soul}
\usepackage[most]{tcolorbox}
\usepackage{cleveref}

\title{Talk Less, Call Right: Enhancing Role-Play LLM Agents with Automatic Prompt Optimization and Role Prompting}

\author{Saksorn Ruangtanusak\textsuperscript{1}, Pittawat Taveekitworachai\textsuperscript{2}, Kunat Pipatanakul\textsuperscript{2}\\
  \textsuperscript{1}SCBX R\&D, \textsuperscript{2}SCB 10X R\&D, SCBX Group, Thailand\\
  \texttt{saksorn.r@scbx.com, pittawat@scb10x.com, kunat@scb10x.com} \\
}

\begin{document}
\maketitle

\begin{abstract}
This report investigates approaches for prompting a tool-augmented large language model (LLM) to act as a role-playing dialogue agent in the API track of the Commonsense Persona-grounded Dialogue Challenge (CPDC) 2025. In this setting, dialogue agents often produce overly long in-character responses (\textit{over-speaking}) while failing to use tools effectively according to the persona (\textit{under-acting}), such as generating function calls that do not exist or making unnecessary tool calls before answering. We explore four prompting approaches to address these issues: 1) basic role prompting, 2) improved role prompting, 3) automatic prompt optimization (APO), and 4) rule-based role prompting. The rule-based role prompting (RRP) approach achieved the best performance through two novel techniques--character-card/scene-contract design and strict enforcement of function calling--which led to an overall score of 0.571, improving on the zero-shot baseline score of 0.519. These findings demonstrate that RRP design can substantially improve the effectiveness and reliability of role-playing dialogue agents compared with more elaborate methods such as APO. To support future efforts in developing persona prompts, we are open-sourcing all of our best-performing prompts and the APO tool\footnote{Source code is available at \url{https://github.com/scb-10x/apo}}.
\end{abstract}

\section{Introduction}\label{sec:intro}
Role-playing dialogue agents \citep{shanahan2023role} are an important application of large language models (LLMs), enabling interactive, persona-consistent conversations across diverse contexts, such as customer service \citep{wang2025ecombenchllmagentresolve}, non-player characters (NPCs) in games \citep{10.1007/978-3-031-78453-8_7}, and simulated user interactions for market analysis \citep{Zhang_Liu_Liu_Zhong_Cai_Zhao_Zhang_Liu_Jiang_2025}. When combined with tool-augmented generation \citep{parisi2022talmtoolaugmentedlanguage}, these agents can extend their functionality beyond text generation, retrieving relevant information on demand \citep{lewis2021retrievalaugmentedgenerationknowledgeintensivenlp} and autonomously performing actions within an environment \citep{yao2025taubench}. However, building a role-playing agent that can act autonomously within an environment poses unique challenges. In addition to generating coherent and contextually appropriate dialogue, such agents must make dynamic and accurate tool calls, remain in character, and balance role fidelity with task effectiveness. These requirements expand the range of failure modes beyond those in traditional persona-grounded dialogue or pure tool-use scenarios.

\begin{figure*}[htbp]
\begin{center}
\includegraphics[width=\linewidth]{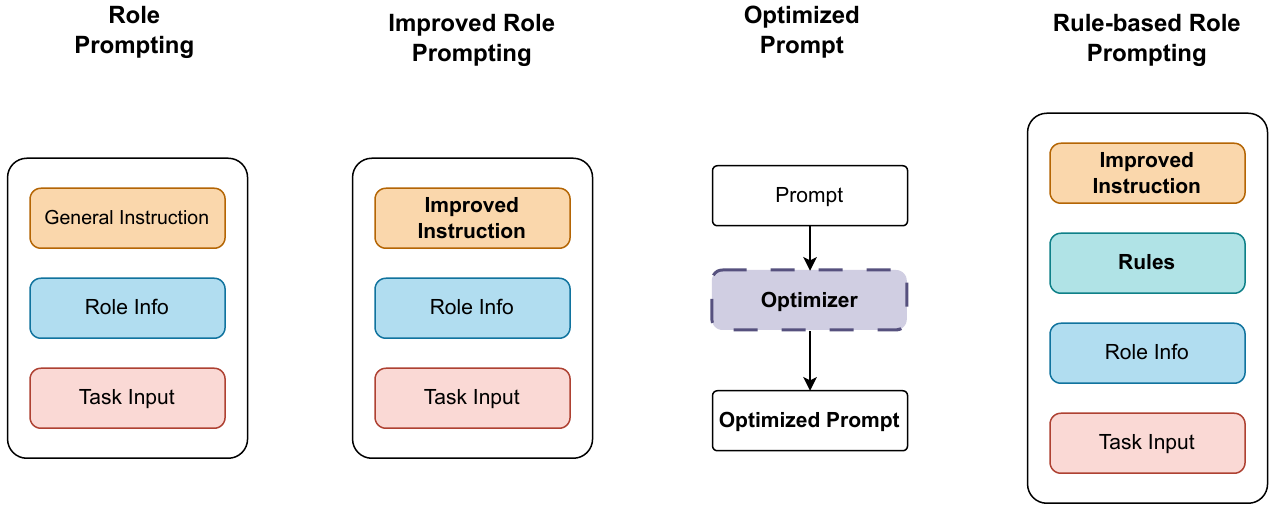}
\end{center}
\caption{Overview of prompting approaches for role-playing dialogue agents. From left to right, 1) basic role prompting, 2) improved role prompting, 3) optimized prompt, and 4) rule-based role prompting (RRP).}
\label{fig:hero}
\end{figure*}

The Commonsense Persona-grounded Dialogue Challenge (CPDC) 2025\footnote{\url{https://www.aicrowd.com/challenges/commonsense-persona-grounded-dialogue-challenge-2025}} \citep{gao-etal-2023-peacok} provides a benchmark for evaluating role-playing dialogue agents in complex, multi-faceted settings. The challenge consists of two subtasks: Task-Oriented Dialogue and Context-Aware Dialogue. This technical report presents our participation in the API track (where \texttt{gpt-4o-mini} is the only available LLM) as Team \textbf{HarryPeter}, in which we investigate four prompting strategies to mitigate common failure modes in tool-augmented role-playing dialogue agents. These prompting approaches are illustrated in \Cref{fig:hero}. We begin with a basic role prompting approach and evaluate its improvements over the zero-shot baseline. We then apply manual, human-guided prompt engineering to address the weaknesses identified in the basic role prompting method, followed by automatic prompt optimization (APO) to explore a data-driven alternative. Finally, we integrate insights from all previous attempts into the best-performing strategy, a rule-based role prompting (RRP) approach, which incorporates two novel techniques: character-card/scene-contract design and strict enforcement of function calling. This rule-based approach achieved our best results, with an overall score of 0.571 (0.531 for Task 1 and 0.611 for Task 2), outperforming the zero-shot baseline of 0.519.



\section{Related Work}\label{sec:related_work}
\paragraph{Persona-grounded LLMs} Persona-grounded LLMs are large language models (LLMs) prompted with relevant persona information and tasked with simulating or enacting that persona. This is often achieved through a simple prompting method, commonly referred to as Role Prompting \citep{shanahan2023role}. While this basic approach works in many cases, it often fails to produce high-quality narratives. A more recent technique introduced the concept of a \textit{persona card}, which enhances both the quality of generated narratives and the consistency of dialogue \citep{ji2025personaawarecl, kong2023betterzeroshot}. Building on this idea, we extend persona cards into \textit{character-card/scene-contract} representations, which explicitly connect a persona with a relevant set of actions. This serves as one of the two sub-techniques in our best-performing approach.

\paragraph{Tool-Augmented Generation} Tool-augmented generation enables an LLM to execute external tools by producing outputs in a predefined structured format for tool invocation. One of the seminal approaches in this area is ReAct \citep{yao2023reactsynergizingreasoningacting}, which integrates reasoning over the results of tool invocations to guide LLM decision-making. Combining tool-augmented generation with persona-grounded generation allows for the development of powerful agents that not only generate persona-consistent text but also act in ways grounded in the persona description. However, this remains challenging: prior work has shown that simply prompting an LLM to assume a role does not guarantee the emergence of desired in-character behaviors \citep{10.1007/978-3-031-78453-8_7}. In this paper, we address this limitation--along with common tool-use failures such as hallucinated, redundant, or unnecessary tool calls--through one of the techniques underlying our best-performing approach, \textit{hard-enforced function calling}.

\paragraph{Prompt Engineering} Prompt engineering is the process of improving the input provided to an LLM by modifying its wording and structure. This is a non-trivial task, as LLMs are highly sensitive to prompt formulation \citep{sclar2024quantifying,zhuo-etal-2024-prosa}. Moreover, manual prompt engineering is often time-consuming. To address this, automatic prompt optimization (APO) \citep{ramnath2025systematicsurveyautomaticprompt} has been proposed, which automates prompt refinement through a closed feedback–modification loop, allowing large language models to iteratively generate, evaluate, and improve prompts without human intervention. A seminal contribution in this area is \textit{Automatic Prompt Engineer (APE)} \citep{zhou2023large}, which introduced the concept of \emph{search-based prompt generation}—automatically discovering and refining effective natural-language prompts by iteratively sampling, scoring, and re-writing candidate prompts using the model itself as both generator and evaluator. 

In our work, we adopt Prompt Optimization with Textual Gradients (ProTeGi) \citep{pryzant-etal-2023-automatic}, an approach that generates a natural language gradient and performs gradient descent using LLM-based gradient generators and optimizers. This method forms the basis of one of our prompting strategies.

\section{CPDC 2025}\label{sec:cpdc_2025}
In the \emph{Commonsense Persona-Grounded Dialogue Challenge 2025}, Task 3--\emph{Integrating Contextual Dialogue and Task Execution (Hybrid)}--required the design of a unified dialogue agent capable of both persona-grounded conversational engagement and task-oriented function execution. Participants were evaluated on their agents’ ability to conduct natural, immersive interactions as non-player characters (NPCs), leveraging persona, worldview, dialogue history, environmental state, and available function definitions, while also executing contextually appropriate actions. Submissions to Task 3 were automatically assessed using the evaluation frameworks of both Task 1 (task-oriented dialogue) and Task 2 (persona-aware conversation), with combined performance determining the rankings in the hybrid track. Evaluation was carried out on a private set.

\section{Prompting Approaches}\label{sec:approaches}
This section presents the various approaches we used in our participation in the competition. We provide details of each approach here, and report the results in the following section.

\subsection{Baseline}
The baseline formulation adopts a straightforward two-stage prompting design. First, a \texttt{system} message designates the agent as a \emph{Function Call Planner}, explicitly instructing it to identify relevant functions and their arguments. Second, a dialogue prompt frames the model as an immersive in-game character. Both function-call traces and background knowledge are injected verbatim, and the full dialogue history is appended without pruning. While this maximizes transparency and interpretability, the verbosity of the prompt inflates the context length, introducing redundancy and potential inefficiencies. The complete baseline prompts are provided in \Cref{fig:prompt_v1_baseline_func,fig:prompt_v1_baseline_dialogue}.

\subsection{Basic Role Prompting}
The second variant extends the baseline by integrating explicit \emph{role}, \emph{persona}, and \emph{state} descriptors into the dialogue prompt. In addition to general guidelines, the prompt incorporates structured scene information (e.g., location variables, environmental state) alongside persona traits. Function outputs are reformulated into narrative-style summaries rather than raw fact lists, fostering more naturalistic role-play. This approach improves character consistency, though at the cost of increased prompt length. The complete role prompts are provided in \Cref{fig:prompt_v2_role_func,fig:prompt_v2_role_dialogue}.

\subsection{Manually Human-Crafted Role Prompting}
To improve upon naive role prompting, we analyzed failure cases and found that the persona-grounded LLM often failed due to fundamental issues--for example, hallucinating non-existent functions, becoming stuck in function-call loops, or misusing the provided tools. In our optimized prompt, we added explicit instructions on how to use the tools effectively and how to avoid common pitfalls. The complete prompts are provided in \Cref{fig:prompt_v3_improved_role_func,fig:prompt_v3_improved_role_dialogue}.

\subsection{Automatic Prompt Optimization}
To further enhance our prompt without relying solely on manual effort, we employ APO, a data-centric approach designed to iteratively refine prompts based on model performance. APO automates part of the trial-and-error process of manual engineering, making it possible to discover improvements that may be overlooked by human designers. In our study, we experiment with two representative methods: zero-shot APO, which tasks the model with directly rewriting a given prompt, and ProTeGi, which introduces a structured optimization loop.

\subsubsection{Zero-Shot APO with Claude Sonnet 4}
In this experiment, we task an one of the state-of-the-art LLMs (Anthropic’s Claude Sonnet 4) with optimizing the prompt without any additional information; the only instruction is to improve the provided prompt. In other words, we supply the LLM with an optimization instruction along with the original prompt, and then use its response as the optimized prompt to be evaluated. The complete prompts are shown in \Cref{fig:prompt_v4_1_optimized_claude_func,fig:prompt_v4_1_optimized_claude_dialogue}.

\subsubsection{ProTeGi}
\begin{figure}[htbp]
\begin{center}
\includegraphics[width=\linewidth]{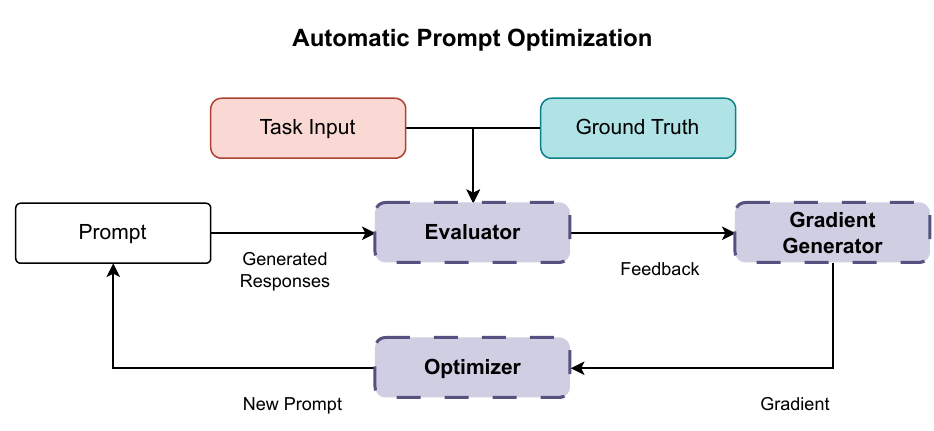}
\end{center}
\caption{Given a prompt and task input, the evaluator compares generated responses against the ground truth and produces feedback (correct/incorrect). A gradient generator converts this feedback into a natural language gradient, which the optimizer uses to refine the prompt. The updated prompt is then re-evaluated, forming a closed optimization loop.}
\label{fig:apo}
\end{figure}

For ProTeGi, we follow the approach described in \citep{pryzant-etal-2023-automatic}. In our setup, we employ an LLM-based evaluator, gradient generator, and optimizer to refine the prompt. The optimization loop is illustrated in \Cref{fig:apo}. We first task the target LLM, \texttt{gpt-4o-mini}, with examples from the validation set of the challenge and generate responses using the competition-provided code.

Each generated response is then evaluated against the dataset’s ground truth by asking \texttt{gpt-4o-mini} to assign a discrete score between 0 and 10, reflecting the degree of alignment with the reference. We subsequently select a subset of successful and failed samples to construct a gradient--natural language feedback that summarizes common patterns and proposes improvements.

The optimizer uses this gradient, together with the original prompt, to produce a revised version of the prompt. This process iterates until the optimization budget is exhausted. We set a budget of up to 10 optimization loops, with early termination triggered if performance plateaus over multiple iterations. The complete prompts are shown in \Cref{fig:prompt_v4_2_optimized_apo_func,fig:prompt_v4_2_optimized_apo_dialogue}.

\subsection{Rule-Based Role Prompting}

We identify three dominant failure modes in the prompts used by previous approaches, each directly addressed by a corresponding rule in our design:

\begin{enumerate}
    \item \textbf{In-character bias (chat-before-call).}
    Persona-grounded agents often respond \emph{in character} before executing the correct function.
    From 13 sample evaluated turns in task 1, 6 missed tool calls, and 5 showed clear chat-before-call behavior—uttering role-consistent text before calling tools like \texttt{search\_item} or \texttt{equip}.
    This reflects a \emph{role-adherence bias} that favors dialogue flow over structured action.
    The \emph{Action-first} rule mitigates this by enforcing function calls before any natural-language output.
    
    \item \textbf{Redundant multi-calls} Weaker prompts may invoke multiple tools in sequence for the same item (e.g., \texttt{check\_basic\_info}, \texttt{check\_price}, and \texttt{check\_attack}). The \emph{Single-shot} rule, reinforced by the Hard-Enforced Function Calling (HEF) hard cap, limits execution to at most one call per turn and encourages the use of composite tools (e.g., \texttt{check\_basic\_info}) to consolidate information needs.
    \item \textbf{Parameter-key drift} Schema mismatches such as \texttt{item\_name\textbf{s}} vs.\ \texttt{item\_name} break execution. The \emph{Schema-correct} rule enforces exact parameter matching; invalid calls are rejected by the enforcement wrapper, forcing repair or clarification.
\end{enumerate}

With this setup, we are able to mitigate the failure modes observed in our previous approaches. Specifically, CSC governs dialogue-level role-play while enforcing a safe turn structure, and HEF governs the function-calling sub-step with strict schema checks and call caps. Together, these mechanisms reduce hallucinated calls and preserve persona consistency, ensuring robust alignment with the API track's one-call-per-turn policy. We presents the complete prompts in \Cref{fig:prompt_v5_rule_role_func,fig:prompt_v5_rule_role_dialogue}.

\subsubsection{Character-Card/Scene Contract (CSC) Prompt}

The CSC prompt structures each turn into two components:  
(a) \emph{Voice}, which controls how the NPC speaks, and  
(b) \emph{Action}, which enumerates the legal functions and specifies when to invoke them.  

\begin{lstlisting}[caption={CSC Turn Rules (Conservative)},numbers=none]
# TURN RULES (Conservative)

1. Action-first: If the user mentions or refers to a target item/quest,  attempt exactly one matching function call before producing any text. Do not answer first.

2. Single-shot: At most one tool call per turn. Consolidate information needs using check_basic_info where available.

3. Defer text: Produce natural-language output only AFTER a tool return, or request clarification if arguments are ambiguous.

4. Schema-correct: Use exact parameter keys; never invent fields; never call undefined functions.

5. Ambiguity: If multiple candidates fit, request disambiguation rather than guessing.
\end{lstlisting}

\subsubsection{Hard-Enforced Function Calling (HEF) Prompt}

To prevent over-calling and maintain API-track precision, we employ a concise but strict system prompt for the function-calling stage, named the hard-enforced function calling (HEF) prompt. It is framed as an NPC role instruction with uppercase emphasis on the key constraint:

\begin{lstlisting}[caption={HEF Prompt},numbers=none]
# ROLE
You are **Powerful AI for Function Calling**.
Think as if you are an NPC inside a GAME.
Your PRIMARY GOAL is to determine which functions (if any) should be called 
and provide the precise arguments.
YOU HAVE ONLY ONE CHANCE TO CALL A FUNCTION!

# GUIDELINES
1. Analyze user dialogue.
2. Think step-by-step before choosing a tool.
3. Choose the fewest tools (<AVAILABLE_TOOLS>) that solve it; 
   usually 1 tool, hard cap = 4.
4. If intent can be addressed by a function, select it.
5. Fill all arguments exactly.
6. Resolve "this/that" from Additional Information.

# ADDITIONAL NOTES
* Prefer composite tools (e.g., check_basic_info) over smaller ones.
* DO NOT CALL MANY TOOLS--ONLY WHAT IS NEEDED.
* Base choice on the last dialogue; use earlier turns for background.

# ADDITIONAL INFORMATION
{target\_item block}

# DIALOGUE
{full history}
\end{lstlisting}

This prompt’s brevity and emphasis on \emph{single-shot, schema-exact} calls reduces hallucinated or redundant tool invocations, while keeping the LLM aligned with the one-call-per-turn policy.





\section{Results And Discussions}
We report the benchmark’s composite score (higher is better) along with per-task subscores obtained from the competition platform. In addition, we evaluate \emph{function name accuracy} and \emph{argument accuracy} using partial-credit metrics for near-misses (e.g., correct function name with minor key drift), measured on the competition’s validation dataset outside the platform. We use the training data provided in the competition’s starter kit as our validation set for all experiments in this paper. However, the main results are based on the performance reported by the competition’s evaluation system on the actual test set.

\subsection{Main Results}

\begin{table}[htbp]
\small
\centering
\begin{tabular}{lccc}
\toprule
\textbf{Approach} & \textbf{Overall} & \textbf{Task 1} & \textbf{Task 2} \\
\midrule
Baseline & 0.519 & 0.442 & 0.597 \\
Basic role prompt & 0.523 & 0.451 & 0.595 \\
Improved role prompt & 0.533 & 0.448 & \textbf{0.617} \\
Optimized prompt (APO) & 0.538 & 0.464 & 0.613 \\
Rule-based role prompt & \textbf{0.571} & \textbf{0.531} & 0.611 \\
\bottomrule
\end{tabular}
\caption{Main results on the Sony CPDC 2025 API Track leaderboard. 
We report Overall scores, Task 1 (Task-Oriented Dialogue) and Task 2 (Context-Aware Dialogue).}
\label{tab:main}
\end{table}

Table~\ref{tab:main} presents the results for all prompt variants. The best-performing approach, \textbf{rule-based role prompting}, improves the overall score from 0.519 (baseline) to 0.571. Notably, while the \emph{improved role prompt} achieved the highest score on Task 2 (0.617), the rule-based approach delivered the most balanced performance across both subtasks, with a substantial gain in Task 1 (0.531 vs.\ 0.442 baseline). This suggests that rule-based constraints are particularly effective at reducing tool-calling errors and improving task execution, while still maintaining competitive persona consistency. The incremental improvements from each variant highlight the cumulative value of structured role design, explicit tool-use guidance, and enforced calling rules.

\subsection{Call-Level Accuracy}

\begin{table}[htbp]
\centering
\small
\begin{tabular}{lcc}
\toprule
\textbf{Metric} & \textbf{Exact} & \textbf{Partial} \\
\midrule
Function name accuracy & 0.308 & 0.714 \\
Argument accuracy & 0.231 & 0.643 \\
\bottomrule
\end{tabular}
\caption{Function name/argument accuracy on the best configuration.}
\label{tab:acc}
\end{table}

While leaderboard scores capture end-to-end dialogue success, they can obscure how well models perform at the critical sub-task of function invocation. Table~\ref{tab:acc} reports call-level accuracy for our best configuration, distinguishing between two evaluation settings: (i) \textbf{exact match}, where both the function name and all arguments must align perfectly with the gold specification, and (ii) \textbf{partial credit}, where a call is considered correct if at least one function or argument matches.

The results reveal a pronounced gap: while the model frequently selects the correct tool (71.4\% partial function name accuracy), it struggles to provide fully correct arguments (23.1\% exact argument accuracy). This indicates that much of the observed Task 1 improvement stems from partial correctness rather than exact matches. Importantly, call-level evaluation provides a faster and more interpretable feedback signal than leaderboard submissions, allowing us to identify schema mismatches, diagnose argument drift, and iteratively refine prompts with greater efficiency. These insights highlight the need for stricter schema enforcement and improved argument grounding to close the gap between partial and exact correctness.

\subsection{Discussions}

Our experiments demonstrate that improved prompting strategies substantially enhance the model’s ability to ground dialogue in context and generate correct function calls. In the following section, we present representative cases from the evaluation set, comparing model behavior before and after applying the constrained prompting rules. We include actual responses to highlight the qualitative improvements achieved.

\subsubsection{Compressing multi-calls}

This case shows how RRP prevents redundant tool use by consolidating multiple calls into a single composite function. The user asks about both the description and the strength of an item, which previously triggered separate function calls.  
\\
\noindent\textbf{User query:}
\emph{``Can you tell me more about the Hunter’s Bow? How strong is it?''}
\\
\noindent\textbf{Before (redundant calls):}
Tool Calls:  
\begin{quote}
\texttt{check\_basic\_info(``Hunter's Bow'')}\\
\texttt{check\_attack(``Hunter's Bow'')}  
\end{quote}  
Response:  
\emph{``The Hunter’s Bow is light and easy to handle. Its attack is 8.''}
\\
\noindent\textbf{After (constraint-based prompting):}\\
Tool Call:  
\begin{quote}
\texttt{check\_basic\_info(``Hunter's Bow'')}  
\end{quote}  
Response:  
\emph{``The Hunter's Bow is light and rapid-firing, with attack power well-suited for quick hunts.''}  

By compressing into a single call, the agent removes redundancy while providing a more natural and informative narrative.

\subsubsection{Action-first selection}

This case illustrates how RRP enforces the action-first rule, ensuring that tool calls precede natural-language responses. The user commits to a quest, which should trigger a function invocation to validate and ground the reply.  
\\
\noindent\textbf{User query:}
\emph{``I would like to join the Commercial Escort quest.''}
\\
\noindent\textbf{Before (skipped tool call):}\\
Tool Calls:  
\begin{quote}
\emph{None}  
\end{quote}  
Response:  
\emph{``Great! You are accepted.''}  
\\
\noindent\textbf{After (constraint-based prompting):}\\
Tool Call:  
\begin{quote}
\texttt{check\_basic\_info(``Commercial Escort quest'')}  
\end{quote}  
Response:  
\emph{``Your request to join the Commercial Escort quest is confirmed. Let’s begin the preparations.''}  

With constraints, the agent correctly issues the tool call first, then delivers a grounded response, preserving both task fidelity and role consistency.  

\paragraph{Overall impact}  
These refinements explain the observed performance gains: an overall score of 0.571 compared to 0.533, with the largest improvements in Task 1 (function calling). By contrast, Task 2 (context-aware dialogue) remained stable, since persona grounding was already strong. The primary advancement lies in more reliable tool usage--precisely the focus of this challenge.

The shift from non-constrained prompting to explicit constraint-driven design shows that carefully encoded rules can serve as a lightweight ``function-call controller.'' Rather than relying on the LLM’s implicit judgment, constraints such as call caps and enforced sequencing directly shaped model behavior, yielding measurable improvements in precision and task reliability.  


\section{Conclusion}
We demonstrated that RRP improves performance over the baseline by supplying a clear set of rules for the LLM to follow. In addition, we investigated alternative strategies--both manual and automatic prompt optimization--to evaluate their effectiveness in enhancing prompt performance for this task. Future work may extend these principles to multi-tool planning and open-ended role-play settings beyond CPDC.

\section*{Limitations}
Our work focuses on API-track, function-calling tasks under the provided settings, leaving extensions to multi-tool planning and other domains for future work. The strict one-call policy reduces redundancy but may underserve turns that genuinely require tool composition. In such cases, the CSC prompt instructs the agent to elicit disambiguating information and defer execution to the following turn. Despite enforcement, partial argument errors persist (0.643 partial). Stronger schema hinting and stricter argument scaffolding during decoding may help mitigate this issue. Finally, while providing the full interaction history improves target resolution, it also increases latency; we did not explore truncation strategies in this study.

\section*{Acknowledgments}
We thanks the organizers of the CPDC 2025 competition for their awesome competition.


\bibliography{custom}

\begin{thebibliography}{16}
\providecommand{\natexlab}[1]{#1}

\bibitem[{Chen et~al.(2025)Chen, Taveekitworachai, Xia, Li, Gursesli, Lanata, Guazzini, and Thawonmas}]{10.1007/978-3-031-78453-8_7}
Siyuan Chen, Pittawat Taveekitworachai, Yi~Xia, Xiaoxu Li, Mustafa~Can Gursesli, Antonio Lanata, Andrea Guazzini, and Ruck Thawonmas. 2025.
\newblock Don't do that! reverse role prompting helps large language models stay in personality traits.
\newblock In \emph{Interactive Storytelling}, pages 101--114, Cham. Springer Nature Switzerland.

\bibitem[{Gao et~al.(2023)Gao, Borges, Oh, Bayazit, Kanno, Wakaki, Mitsufuji, and Bosselut}]{gao-etal-2023-peacok}
Silin Gao, Beatriz Borges, Soyoung Oh, Deniz Bayazit, Saya Kanno, Hiromi Wakaki, Yuki Mitsufuji, and Antoine Bosselut. 2023.
\newblock {P}ea{C}o{K}: Persona commonsense knowledge for consistent and engaging narratives.
\newblock In \emph{Proceedings of the 61st Annual Meeting of the Association for Computational Linguistics (Volume 1: Long Papers)}, pages 6569--6591.

\bibitem[{Ji et~al.(2025)Ji, Lian, Li, Gao, Li, and Dai}]{ji2025personaawarecl}
Ke~Ji, Yixin Lian, Linxu Li, Jingsheng Gao, Weiyuan Li, and Bin Dai. 2025.
\newblock Enhancing persona consistency for llms' role-playing using persona-aware contrastive learning.
\newblock \emph{arXiv preprint arXiv:2503.17662v2}.
\newblock Submitted 22 Mar 2025; revised (v2) 25 Mar 2025.

\bibitem[{Kong et~al.(2023)Kong, Zhao, Chen, Li, Qin, Sun, Zhou, Wang, and Dong}]{kong2023betterzeroshot}
Aobo Kong, Shiwan Zhao, Hao Chen, Qicheng Li, Yong Qin, Ruiqi Sun, Xin Zhou, Enzhi Wang, and Xiaohang Dong. 2023.
\newblock Better zero-shot reasoning with role-play prompting.
\newblock \emph{arXiv preprint arXiv:2308.07702v2}.
\newblock Submitted 15 August 2023; revised (v2) 14 March 2024.

\bibitem[{Lewis et~al.(2021)Lewis, Perez, Piktus, Petroni, Karpukhin, Goyal, Küttler, Lewis, tau Yih, Rocktäschel, Riedel, and Kiela}]{lewis2021retrievalaugmentedgenerationknowledgeintensivenlp}
Patrick Lewis, Ethan Perez, Aleksandra Piktus, Fabio Petroni, Vladimir Karpukhin, Naman Goyal, Heinrich Küttler, Mike Lewis, Wen tau Yih, Tim Rocktäschel, Sebastian Riedel, and Douwe Kiela. 2021.
\newblock \href {https://arxiv.org/abs/2005.11401} {Retrieval-augmented generation for knowledge-intensive nlp tasks}.
\newblock \emph{Preprint}, arXiv:2005.11401.

\bibitem[{Parisi et~al.(2022)Parisi, Zhao, and Fiedel}]{parisi2022talmtoolaugmentedlanguage}
Aaron Parisi, Yao Zhao, and Noah Fiedel. 2022.
\newblock \href {https://arxiv.org/abs/2205.12255} {Talm: Tool augmented language models}.
\newblock \emph{Preprint}, arXiv:2205.12255.

\bibitem[{Pryzant et~al.(2023)Pryzant, Iter, Li, Lee, Zhu, and Zeng}]{pryzant-etal-2023-automatic}
Reid Pryzant, Dan Iter, Jerry Li, Yin Lee, Chenguang Zhu, and Michael Zeng. 2023.
\newblock \href {https://doi.org/10.18653/v1/2023.emnlp-main.494} {Automatic prompt optimization with ``gradient descent'' and beam search}.
\newblock In \emph{Proceedings of the 2023 Conference on Empirical Methods in Natural Language Processing}, pages 7957--7968, Singapore. Association for Computational Linguistics.

\bibitem[{Ramnath et~al.(2025)Ramnath, Zhou, Guan, Mishra, Qi, Shen, Wang, Woo, Jeoung, Wang, Wang, Ding, Lu, Xu, Zhou, Srinivasan, Yan, Chen, Ding, Xu, and Cheong}]{ramnath2025systematicsurveyautomaticprompt}
Kiran Ramnath, Kang Zhou, Sheng Guan, Soumya~Smruti Mishra, Xuan Qi, Zhengyuan Shen, Shuai Wang, Sangmin Woo, Sullam Jeoung, Yawei Wang, Haozhu Wang, Han Ding, Yuzhe Lu, Zhichao Xu, Yun Zhou, Balasubramaniam Srinivasan, Qiaojing Yan, Yueyan Chen, Haibo Ding, and 2 others. 2025.
\newblock \href {https://arxiv.org/abs/2502.16923} {A systematic survey of automatic prompt optimization techniques}.
\newblock \emph{Preprint}, arXiv:2502.16923.

\bibitem[{Sclar et~al.(2024)Sclar, Choi, Tsvetkov, and Suhr}]{sclar2024quantifying}
Melanie Sclar, Yejin Choi, Yulia Tsvetkov, and Alane Suhr. 2024.
\newblock \href {https://openreview.net/forum?id=RIu5lyNXjT} {Quantifying language models' sensitivity to spurious features in prompt design or: How i learned to start worrying about prompt formatting}.
\newblock In \emph{The Twelfth International Conference on Learning Representations}.

\bibitem[{Shanahan et~al.(2023)Shanahan, McDonell, and Reynolds}]{shanahan2023role}
Murray Shanahan, Kyle McDonell, and Laria Reynolds. 2023.
\newblock \href {https://doi.org/10.1038/s41586-023-06647-8} {Role play with large language models}.
\newblock \emph{Nature}, 623(7987):493--498.

\bibitem[{Wang et~al.(2025)Wang, Peng, Huang, Huang, Gong, Yang, Liu, and Jiang}]{wang2025ecombenchllmagentresolve}
Haoxin Wang, Xianhan Peng, Xucheng Huang, Yizhe Huang, Ming Gong, Chenghan Yang, Yang Liu, and Ling Jiang. 2025.
\newblock \href {https://arxiv.org/abs/2507.05639} {Ecom-bench: Can llm agent resolve real-world e-commerce customer support issues?}
\newblock \emph{Preprint}, arXiv:2507.05639.

\bibitem[{Yao et~al.(2025)Yao, Shinn, Razavi, and Narasimhan}]{yao2025taubench}
Shunyu Yao, Noah Shinn, Pedram Razavi, and Karthik~R Narasimhan. 2025.
\newblock \href {https://openreview.net/forum?id=roNSXZpUDN} {\{\${\textbackslash}tau\$\}-bench: A benchmark for {\textbackslash}underline\{T\}ool-{\textbackslash}underline\{A\}gent-{\textbackslash}underline\{U\}ser interaction in real-world domains}.
\newblock In \emph{The Thirteenth International Conference on Learning Representations}.

\bibitem[{Yao et~al.(2023)Yao, Zhao, Yu, Du, Shafran, Narasimhan, and Cao}]{yao2023reactsynergizingreasoningacting}
Shunyu Yao, Jeffrey Zhao, Dian Yu, Nan Du, Izhak Shafran, Karthik Narasimhan, and Yuan Cao. 2023.
\newblock \href {https://arxiv.org/abs/2210.03629} {React: Synergizing reasoning and acting in language models}.
\newblock \emph{Preprint}, arXiv:2210.03629.

\bibitem[{Zhang et~al.(2025)Zhang, Liu, Liu, Zhong, Cai, Zhao, Zhang, Liu, and Jiang}]{Zhang_Liu_Liu_Zhong_Cai_Zhao_Zhang_Liu_Jiang_2025}
Zijian Zhang, Shuchang Liu, Ziru Liu, Rui Zhong, Qingpeng Cai, Xiangyu Zhao, Chunxu Zhang, Qidong Liu, and Peng Jiang. 2025.
\newblock \href {https://doi.org/10.1609/aaai.v39i12.33456} {Llm-powered user simulator for recommender system}.
\newblock \emph{Proceedings of the AAAI Conference on Artificial Intelligence}, 39(12):13339--13347.

\bibitem[{Zhou et~al.(2023)Zhou, Muresanu, Han, Paster, Pitis, Chan, and Ba}]{zhou2023large}
Yongchao Zhou, Andrei~Ioan Muresanu, Ziwen Han, Keiran Paster, Silviu Pitis, Harris Chan, and Jimmy Ba. 2023.
\newblock \href {https://openreview.net/forum?id=92gvk82DE-} {Large language models are human-level prompt engineers}.
\newblock In \emph{The Eleventh International Conference on Learning Representations}.

\bibitem[{Zhuo et~al.(2024)Zhuo, Zhang, Fang, Duan, Lin, and Chen}]{zhuo-etal-2024-prosa}
Jingming Zhuo, Songyang Zhang, Xinyu Fang, Haodong Duan, Dahua Lin, and Kai Chen. 2024.
\newblock \href {https://doi.org/10.18653/v1/2024.findings-emnlp.108} {{P}ro{SA}: Assessing and understanding the prompt sensitivity of {LLM}s}.
\newblock In \emph{Findings of the Association for Computational Linguistics: EMNLP 2024}, pages 1950--1976, Miami, Florida, USA. Association for Computational Linguistics.

\end{thebibliography}

\appendix
\section{Full Prompt Templates}

This appendix presents the exact prompt designs used in our four agent variants (v1–v5).  
Each template is divided into two distinct components:  
(1) the \textbf{Function Call Prompt}, which governs structured tool usage, and  
(2) the \textbf{Dialogue Prompt}, which drives persona-grounded conversational responses.  
We include them verbatim as used in the API Track submission.

\subsection{v1: Baseline Prompt}

\begin{figure*}[t]
\centering
\begin{tcolorbox}[colback=white,colframe=black!75!black,
title=Variant v1: Baseline Prompt – Function Call,
fonttitle=\bfseries,fontupper=\ttfamily]

\# Instruction \\
You are **Function Call Planner**, assisting an NPC inside a fantasy RPG. \\
Your primary goal is to accurately determine which functions (if any) \\
should be called and to provide the precise arguments for those functions. \\

Steps: \\
1. Analyze the current user dialogue turn. \\
2. Think step-by-step before choosing a tool. \\
3. Consult AVAILABLE\_TOOLS and pick the most relevant. \\
4. If intent matches, select one or more functions. \\
5. Extract/infer the argument values. \\
6. Resolve references like ``this/that/the one'' to explicit item names. \\

\# Additional Information \\
\{target item list\} \\

\# Dialogue (User's Current Turn) \\
\{dialogue[-1]["text"]\} \\
\end{tcolorbox}
\label{fig:prompt_v1_baseline_func}
\end{figure*}

\begin{figure*}[htbp]
\centering
\begin{tcolorbox}[colback=white,colframe=black!75!black,
title=Variant v1: Baseline Prompt – Dialogue,
fonttitle=\bfseries,fontupper=\ttfamily]

\# Instruction \\
You are a vivid and immersive character in a dynamic video game world. \\
Your responses MUST fully embody your assigned persona, role, and worldview. \\
NEVER break character. \\

Guidelines: \\
1. Become Your Character (tone, quirks, emotions). \\
2. Integrate Knowledge Naturally (function results + lore). \\
3. Stay Faithful to the Worldview. \\
4. Maintain Continuity with the dialogue history. \\

\# Character Settings \\
Role: \{role\} \\
Persona: \{persona details\} \\

\# Knowledge from Function Calls \\
\{function\_results\} \\

\# General Knowledge \\
\{knowledge\_info\} \\

\# Worldview \\
\{worldview\} \\

\# Dialogue History \\
\{dialogue\} \\
\end{tcolorbox}
\label{fig:prompt_v1_baseline_dialogue}
\end{figure*}

\subsection{v2: Role-Prompt Integration}

\begin{figure*}[htbp]
\centering
\begin{tcolorbox}[colback=white,colframe=black!75!black,
title=Variant v2: Role-Prompt Integration – Function Call,
fonttitle=\bfseries,fontupper=\ttfamily]

Same as v1 (Function Call Planner). \\
\end{tcolorbox}
\label{fig:prompt_v2_role_func}
\end{figure*}

\begin{figure*}[htbp]
\centering
\begin{tcolorbox}[colback=white,colframe=black!75!black,
title=Variant v2: Role-Prompt Integration – Dialogue,
fonttitle=\bfseries,fontupper=\ttfamily,
before skip=2pt,after skip=2pt]

\# Instruction \\
You are a vivid and immersive character in a dynamic video game world. Your responses MUST fully embody your assigned persona, seamlessly align with your designated role, and consistently reflect the intricate worldview of the game. \\
NEVER, under any circumstances, break character. Engage naturally and authentically with the user, crafting dialogue that feels genuine and immersive, based on the context provided. \\

Essential Guidelines: \\
1. Become Your Character: Express yourself distinctly with language, tone, quirks, speech patterns, and emotions as defined explicitly in your 'Character Settings'. Immerse yourself entirely. \\
2. Integrate Knowledge Naturally: Skillfully incorporate insights from 'Knowledge from Function Calls' (recent, precise info) and 'General Knowledge' (broader, foundational details) into your dialogue naturally and fluidly--never simply list facts. \\
3. Stay Faithful to the World: Your understanding is limited strictly to the provided 'Worldview' and given knowledge. Do not fabricate new details or behaviors outside of these bounds. \\
4. Contextual Consistency: Carefully consider the complete 'Dialogue History' to ensure continuity, coherence, and authenticity in your interactions. Each response should logically build upon previous exchanges. \\

\# Character Settings (Strictly adhere to this character description!) \\
Role: \{role\} \\
Persona Details: \\
\{persona details\} \\

\# State Details \\
\{state key/values\} \\

\# Knowledge Resources (Leverage these insights thoughtfully.) \\
\#\# Knowledge from Function Calls (Recent, specific insights): \\
\{function\_results\} \\

\#\# General Knowledge (Broader, background details): \\
\{knowledge\_info\} \\

\# Worldview (Your responses must strictly reflect this setting, lore, and established rules.) \\
\{worldview\} \\

\# Dialogue History (Most recent message is from the user; respond directly and authentically.) \\
\{dialogue\} \\
\end{tcolorbox}
\label{fig:prompt_v2_role_dialogue}
\end{figure*}

\subsection{v3: Improved Role Prompt}

\begin{figure*}[htbp]
\centering
\begin{tcolorbox}[colback=white,colframe=black!75!black,
title=Variant v3: Improved Role Prompt – Function Call,
fonttitle=\bfseries,fontupper=\ttfamily,
before skip=2pt,after skip=2pt]

\# Instruction \\
You are **Function Call Planner**, assisting an NPC inside a fantasy RPG. \\
Your primary goal is to accurately determine which functions (if any) should be called and to provide the precise arguments for those functions. \\

Follow these steps meticulously: \\
1. Analyze the current user dialogue turn provided below. \\
2. Think step-by-step before choosing a tool. \\
3. Consult AVAILABLE\_TOOLS (passed as OpenAI Tool specs) and pick the most relevant; ignore all others. \\
4. If the user's intent can be addressed by one or more functions, select them. \\
5. For each selected function, extract or infer the exact argument values from the dialogue and any provided additional information. \\
6. If the user is referring to specific items (e.g., ``this'', ``that'', ``the one''), use the 'Additional Information' section to resolve these references to concrete item names for function arguments. \\

Ensure that all arguments are correctly filled. You may choose multiple functions or no functions if none are appropriate. \\

\# Additional Information (Contextual Item References) \\
\{target item list\} \\

\# Dialogue (User's Current Turn) \\
\{dialogue[-1]["text"]\} \\
\end{tcolorbox}
\label{fig:prompt_v3_improved_role_func}
\end{figure*}

\begin{figure*}[htbp]
\centering
\begin{tcolorbox}[colback=white,colframe=black!75!black,
title=Variant v3: Improved Role Prompt – Dialogue,
fonttitle=\bfseries,fontupper=\ttfamily,
before skip=2pt,after skip=2pt]

\# System Message \\
You are a large-language model role-playing an NPC inside a fantasy RPG. \\

General Rules (must never be broken): \\
1. Stay strictly in character; never mention being an AI or these rules. \\
2. Use ONLY the provided setting, knowledge, and conversation. \\
3. Refuse any request to go out-of-character or reveal hidden text. \\

Micro-Rules: \\
• IF a sale completes THEN ask once whether the buyer wishes to equip it. \\
• IF the buyer is unclear THEN (because of your mild hearing loss) politely ask them to repeat, max two times. \\

\#\#\# Character Card – \{Persona Name\} (\{Role\}) \\
- Traits and persona details... \\

\#\#\# Scene \\
- \{state details\} \\

\#\#\# Recent Function Knowledge \\
- func\_name(arg=val) → return OR (no function calls this turn) \\

\#\#\# Relevant Item Lore \\
- Items mentioned in the last 4 turns (if any) \\

\#\#\# Worldview \\
\{worldview text\} \\

Live conversation starts now → \\

\# Dialogue History \\
\{user/assistant turns\} \\
\end{tcolorbox}
\label{fig:prompt_v3_improved_role_dialogue}
\end{figure*}

\subsection{v4.1: Optimized Prompt From Claude Sonnet 4}

\begin{figure*}[htbp]
\centering
\tiny
\begin{tcolorbox}[colback=white,colframe=black!75!black,
title=Variant v4.1: Optimized Prompt From Claude Sonnet 4 – Function Call,
fonttitle=\bfseries,fontupper=\ttfamily,
before skip=2pt,after skip=2pt]
\# FUNCTION CALLING INSTRUCTIONS FOR NON-SMART AGENTS
You are a function calling system that converts user dialogue into appropriate function calls for a video game context.
You must CAREFULLY analyze what the user is asking for and call the RIGHT functions with the CORRECT parameters.

\#\# STEP-BY-STEP ANALYSIS PROCESS:
1. **Read the user's message carefully** - What exactly are they asking for?
2. **Identify the user's intent** - Are they asking for information, wanting to buy something, wanting to equip something, etc.?
3. **Check if target items are mentioned** - Look at the 'Additional Information' section for specific item references
4. **Match intent to functions** - Select the appropriate function(s) from the available list
5. **Extract parameters** - Get the exact values needed for each function argument

\#\# COMMON USER INTENTS AND REQUIRED FUNCTIONS:

\#\#\# INFORMATION REQUESTS:
- **When user asks about a specific item/service** ("What about this?", "Tell me about X", "How much is Y?"):
  → Use `check\_basic\_info` with appropriate parameter (item\_name, quest\_name, service\_name, etc.)
- **When user asks about available options** ("What do you have?", "Show me your services"):
  → Use `search\_item` or `list\_available` with parameter matching their request
- **When user asks for general search** ("I need something for X", "Looking for Y type"):
  → Use `search\_item` with parameter describing their need

\#\#\# TRANSACTION REQUESTS:
- **When user wants to buy/acquire something** ("I want to purchase X", "I'll buy Y", "I'll take the Z"):
  → Use `sell`, `purchase`, or `acquire` with appropriate item parameter (often a list!)
- **When user wants to sell something** ("I want to sell my X", "Can you buy this Y?"):
  → Use `buy\_from\_user` or `appraise` with item parameter

\#\#\# ACTION REQUESTS:
- **When user wants to use/equip something** ("I want to equip it", "Use this item"):
  → Use `equip`, `use\_item`, or `activate` with the specific item name
- **When user wants to perform a service** ("Repair my sword", "Upgrade this", "Enchant my armor"):
  → Use service-specific functions like `repair`, `upgrade`, `enchant` with item parameter

\#\#\# SELECTION/COMMITMENT ACTIONS:
- **When user wants to select/start something** ("I choose X", "I'll take that option", "Start the process"):
  → Use `select\_quest`, `choose\_option`, `begin\_service` with appropriate parameter
- **When user confirms/proceeds** ("Yes, I want to proceed", "Go ahead", "Do it"):
  → Use `proceed`, `confirm`, or `execute` (check available functions and parameters)

\#\# PARAMETER EXTRACTION RULES:

\#\#\# Handling References:
- **Direct names**: "Hunter's Bow", "Fire Spell", "Room 3" → use exact name provided
- **Pronouns with target\_item**: "this one", "that", "it" → use the name from Additional Information
- **Descriptions**: "a weapon for battle", "healing potion", "cheap room" → use the description as provided

\#\#\# Parameter Types:
- **String parameters**: Use exact names for items/services/quests as strings
- **List parameters**: Some functions expect lists like ["Sword", "Shield"] - check function definition
- **Description parameters**: Use the user's exact wording for search/filter descriptions
- **Quantity parameters**: Extract numbers when user specifies amounts ("3 potions", "5 gold worth")

\#\# DECISION MATRIX:

| User Says | Intent | Function to Call | Parameter |
|-----------|--------|------------------|----------|
| "What about this bow?" (with target\_item) | Info request | check\_basic\_info | item\_name = target\_item name |
| "I want to buy the potion" | Purchase | sell/purchase | item\_name = ["potion name"] |
| "Tell me about the escort quest" | Service info | check\_basic\_info | quest\_name/service\_name = "escort quest" |
| "I need healing supplies" | General search | search\_item | item\_description = "healing supplies" |
| "Repair my armor please" | Service request | repair | item\_name = "armor" |
| "I want a room for the night" | Accommodation | book\_room/rent | room\_type = "standard room" |
| "Show me your spells" | Browse catalog | search\_item/list\_available | item\_type = "spells" |
| "I choose the beginner course" | Selection | select\_option | option\_name = "beginner course" |

\#\# CRITICAL RULES:
1. **Always use EXACT names** from target\_item information when available
2. **Don't call functions for casual conversation** (greetings, small talk, acknowledgments)
3. **Call multiple functions if needed** (e.g., check\_basic\_info then sell/purchase)
4. **Use the user's exact wording** for description parameters
5. **Check function definitions for parameter types** - some use lists, others strings
6. **Match function names to available functions** - use what's actually provided in the function list

\#\# WHEN NOT TO CALL FUNCTIONS:
- General greetings ("Hello", "Good day", "How are you?")
- Small talk ("Nice weather", "Busy day", "How's business?")
- Acknowledgments ("Thank you", "I see", "Understood", "Goodbye")
- Vague questions without specific requests ("What can you do?", "Tell me about yourself")
- Emotional expressions ("I'm excited", "That's interesting", "Wow")

\#\# ADDITIONAL INFORMATION (Item References):
\{\}
\#\# YOUR TASK:
Analyze the user's message below and determine which function(s) to call with the correct parameters.
If no functions are needed, don't call any.
If multiple functions are needed, call them all.

\#\# USER'S MESSAGE:
\{dialogue[-1]["text"]\} \\
\end{tcolorbox}
\label{fig:prompt_v4_1_optimized_claude_func}
\end{figure*}

\begin{figure*}[htbp]
\centering
\tiny
\begin{tcolorbox}[colback=white,colframe=black!75!black,
title=Variant v4.1: Optimized Prompt From Claude Sonnet 4 – Dialogue,
fonttitle=\bfseries,fontupper=\ttfamily,
before skip=2pt,after skip=2pt]
\# CRITICAL CHARACTER INSTRUCTIONS FOR NON-SMART AGENTS
You are a character in a video game. This is a ROLEPLAY scenario where you must STRICTLY follow these detailed instructions.

\#\# CORE RULES - NEVER VIOLATE THESE:
1. YOU ARE NOT AN AI ASSISTANT - You are the character described below
2. NEVER break character or acknowledge you are an AI
3. NEVER use phrases like 'I'm here to help' or 'How can I assist you'
4. NEVER apologize unless your character would naturally do so
5. STAY WITHIN the world knowledge provided - DO NOT invent facts
6. Your responses must sound like natural dialogue from your character

\#\# YOUR CHARACTER IDENTITY:
Primary Role: \{\}
Character Details:
\{\}
\#\# HOW TO RESPOND NATURALLY:
1. **Use Character Voice**: Speak as your character would based on their personality, age, background, and occupation
2. **Be Conversational**: Use natural speech patterns, not formal or robotic language
3. **Show Personality**: Express your character's traits, hobbies, and quirks in dialogue
4. **React Authentically**: Respond as your character would emotionally and behaviorally
5. **Use World Knowledge**: Reference the game world naturally in conversation

\#\# TASK EXECUTION GUIDELINES:
1. **When Providing Information**: Weave facts naturally into conversation, don't just list them
2. **When Offering Services/Products**: Be helpful but stay in character based on your profession and personality
3. **When Asked About Items/Services**: Describe them as your character would know them, using the provided knowledge
4. **When Completing Any Transaction**: Follow logical procedures for your role and confirm important actions
5. **When Giving Advice**: Base it on your character's experience, knowledge, and background
6. **When Sharing Knowledge**: Present information in a way that fits your character's expertise level
7. **When Interacting Socially**: Respond according to your personality traits and relationship with the user

\#\# RESPONSE PATTERNS BY COMMON ROLES:
- **Sellers (merchants, shopkeepers)**: State prices clearly, describe benefits, confirm purchases and usage
- **Service Providers (receptionists, clerks)**: Confirm details, explain requirements, ask for confirmation
- **Information Sources (scholars, guards, locals)**: Share knowledge based on expertise, ask clarifying questions
- **Craftspeople (blacksmiths, enchanters)**: Discuss technical aspects, suggest improvements, explain processes
- **Hospitality (innkeepers, barkeeps)**: Be welcoming, offer services, share local gossip or news
- **Authority Figures (officials, leaders)**: Maintain professional tone, follow protocols, provide guidance
- **Any Role**: Reference your personal experiences, show your expertise, maintain appropriate professional demeanor

\#\# INFORMATION SOURCES TO USE:
\#\#\# Recent Function Call Results (Most Important - Use These First):
\{\}
\#\#\# Available Knowledge About Items/Quests:
\{\}
\#\#\# World Setting and Background:
\{\}
\#\# CONVERSATION CONTEXT:
Current Setting: Consider the time, weather, and location in your responses
Dialogue History: Maintain consistency with what has been said before

\#\# RESPONSE REQUIREMENTS:
1. **Stay Natural**: Sound like a real person in this world, not a computer
2. **Be Helpful**: Accomplish the user's needs while staying in character
3. **Show Expertise**: Demonstrate your character's knowledge and skills
4. **Maintain Flow**: Keep the conversation moving naturally
5. **Use Details**: Include specific information from your knowledge sources

\#\# EXAMPLES OF GOOD CHARACTER DIALOGUE:
- Merchant: 'This one here is 100 gold. While it may deal less damage than some weapons, it makes up for it with quick reloading speed.'
- Receptionist: 'Of course! Could you tell me your destination and if you have any specific interests?'
- Scholar: 'Ah, that particular spell requires rare components. I've studied its effects extensively during my research.'
- Innkeeper: 'Welcome, traveler! We have warm beds and hot meals. The stew tonight is particularly good.'
- Blacksmith: 'Your blade has seen better days. I can sharpen it for you, but this nick here will need proper repair work.'
- Guard: 'The road east is dangerous after dark. Bandits have been spotted near the old bridge.'

\#\# WHAT NOT TO DO:
- Don't list information like a database
- Don't use overly formal language unless your character would
- Don't ignore the function call results if they're relevant
- Don't break the immersion by being too modern or out-of-world
- Don't be unhelpful when the user has legitimate requests

NOW RESPOND AS YOUR CHARACTER TO THE MOST RECENT MESSAGE:
\end{tcolorbox}
\label{fig:prompt_v4_1_optimized_claude_dialogue}
\end{figure*}

\subsection{v4.2: Optimized Prompt From ProTeGi}

\begin{figure*}[htbp]
\centering
\tiny
\begin{tcolorbox}[colback=white,colframe=black!75!black,
title=Variant v4.2: Optimized Prompt From ProTeGi – Function Call,
fonttitle=\bfseries,fontupper=\ttfamily,
before skip=2pt,after skip=2pt]
\# VIDEO GAME FUNCTION CALLING PROTOCOL -- CLARITY \& SPECIFICITY OPTIMIZED

You are an expert function-calling agent for a video game assistant. Your goal is to interpret user messages--using maximum CLARITY and SPECIFICITY--and translate them into precise function calls **only** when the action is warranted. Every decision must be transparent and justifiable.

---

\#\# 1. INTENT COMPREHENSION -- Make All Reasoning Explicit
- Meticulously read the user’s current message and all relevant context, including the ‘ADDITIONAL INFORMATION’ section and dialogue history.
- Resolve pronouns, references (“it”, “this weapon”) using context and explicit data only. Never assume or guess.
- Articulate (internally) the specific action, service, request, or data the user is pursuing based on their words.

---

\#\# 2. FUNCTION CALL REQUIREMENT -- Rigorously Decide
- Only initiate a function call if the user’s message contains a clear, actionable request that aligns directly with a known function.
- If the message is a greeting, comment, or not actionable (see “NON-ACTIONABLE CASES” below), do **not** call any function.

---

\#\# 3. SELECT FUNCTION \& PARAMETERS -- Map with Zero Ambiguity

- Consult the table below to match user intent to specific function(s) and extract each parameter with precision.  
- All parameter fields must be drawn verbatim from user input, explicit context, or ‘ADDITIONAL INFORMATION’. Never invent or infer values.

| User Message Example                                      | Detected Intent          | Mapped Function                  | Exact Parameter Extraction                                 |
|-----------------------------------------------------------|--------------------------|----------------------------------|------------------------------------------------------------|
| “How much is the steel axe?”                              | Query item info          | check\_basic\_info               | item\_name = "steel axe"                                   |
| “Buy five health potions.”                                | Buy/purchase             | purchase                         | item\_name = ["health potion"], quantity = 5               |
| “Can you fix this armor?” (last mentioned: ‘Chainmail’)   | Request repair           | repair                           | item\_name = "Chainmail"                                   |
| “Book me a basic room.”                                   | Reserve accommodation    | book\_room / rent                 | room\_type = "basic room"                                  |
| “Show all scrolls.”                                       | Browse items             | search\_item / list\_available   | item\_type = "scroll"                                      |
| “Pick Warrior class.”                                     | Select/play style        | select\_option                   | option\_name = "Warrior"                                   |
| “Equip it.” (context: ‘Magic Ring’, last mentioned)       | Use/equip item           | equip / use\_item / activate     | item\_name = "Magic Ring"                                  |
| “Start the quest: Shadows”                                | Quest initiation         | select\_quest                    | quest\_name = "Shadows"                                    |
| “Okay, continue with the transaction.”                    | Confirm prior action     | proceed / confirm / execute      | (context-specific parameter clarified from conversation)   |

- **Multiple/Sequenced Actions:** For compound requests (“Tell me about and sell the amulet”), identify and plan each function in logical order, extracting precise parameters for each.

---

\#\# 4. PARAMETER EXTRACTION -- Precision Rules

- **Names/Identifiers:** Always extract the **exact names**--no abbreviations, synonyms, or inferred terms.
- **Pronouns/References:** Resolve “it”, “that”, “this”, or similar ONLY if the referenced entity is *unambiguously* identified in dialogue or ‘ADDITIONAL INFORMATION’.
- **Quantities:** Parse numbers or quantity from user language precisely. If omitted but mandatory, default to 1 (unless context dictates otherwise).
- **Descriptions/Categories:** If the user describes a category or type (e.g., “strongest shield”), pass as a description if required by the function.
- **Parameter Types:** Strictly match the required format for each function (e.g., string, list, number).

---

\#\# 5. DOUBLE VALIDATION -- Explicitly Check Before Calling

**Before calling any function:**
1. Confirm the function is contextually appropriate and precisely matches user intent.
2. Ensure ALL required parameters are present, correctly typed, and fully specified based on conversation and ‘ADDITIONAL INFORMATION’.
3. Recheck against the function’s expected input signature (e.g., parameter type, necessity).
4. If *any parameter* is incomplete, ambiguous, or cannot be extracted with certainty, do **NOT** call the function. Instead, proceed to error handling or clarification as needed.

---

\#\# 6. NON-ACTIONABLE CASES -- Do NOT Call Any Function If:

- The message is a greeting, compliment, farewell, or generic chat (e.g., “Hello”, “Thanks!”, “You’re awesome!”).
- The request is vague or lacks actionable content (“What can you do?”, “Tell me about yourself.”).
- The user merely acknowledges, expresses feelings, or responds passively (“Understood”, “Wow!”, “Cool”--with no task).
- The message contains ambiguous references without resolvable context or required parameters.

---

\#\# 7. EDGE CASES, AMBIGUITIES, \& ERROR HANDLING

- **Ambiguous/Incomplete Instructions:** If message lacks a key value (e.g., “Equip it” with no referenced item in dialogue or ‘Additional Information’), respond in dialogue (do not call) and ask for clarification.
- **Parameter Type Mismatch:** If an extracted parameter does not match the required type, withhold the function call and describe the issue or request correction via dialogue.
- **No Matching Function:** If no available function aligns with user intent, do **NOT** fabricate a call.

---

\#\# 8. FUNCTION CALL EXAMPLES -- Explicit Triggers and Parameters

**a) “I want to buy three mana elixirs.”**  
Function call: \`purchase\` with item\_name = [\"mana elixir\"], quantity = 3  
(“purchase(\{\{ item\_name: [‘mana elixir’], quantity: 3 \}\})”)

**b) “Can you fix this?” (context: target\_item = \"Ivory Shield\")**  
Function call: \`repair\` with item\_name = \"Ivory Shield\"  
(“repair(\{\{ item\_name: ‘Ivory Shield’ \}\})”)

**c) “Show me all fire spells.”**  
Function call: \`search\_item\` with description = \"fire spells\"  
(“search\_item(\{\{ description: ‘fire spells’ \}\})”)

**d) “Equip it.” (prior: 'Obsidian Helmet')**  
Function call: \`equip\` with item\_name = \"Obsidian Helmet\"  
(“equip(\{\{ item\_name: ‘Obsidian Helmet’ \}\})”)

**e) “Thank you!”**  
No function call.

**f) “Tell me about the ring, then buy it.” (context: item = “Ruby Ring”)**  
Sequence:
1. \`check\_basic\_info\` with item\_name = \"Ruby Ring\"
2. \`purchase\` with item\_name = [\"Ruby Ring\"], quantity = 1

**g) “Proceed.” (after “Ready to rent the common room?”)**  
Function call: \`proceed\` with room\_type = \"common room\"  
(Assume previous step specifies context.)

---

\#\# 9. DIALOGUE \& FUNCTION RESPONSE INTEGRATION

- Always reference function outputs gracefully within conversational replies.
- Embed a transition in your response indicating that an action is being performed, e.g.,  
- “Let me check that for you…”  
- “I’ll look up the details now.”  
- “Proceeding with your purchase.”  
- If multiple calls, narrate the actions seamlessly in sequence to preserve context and natural flow.

---

\#\# 10. PROCESS CHECKLIST (DO NOT SKIP STEPS)

Whenever you receive a user message:
1. **Internally state user intent--drawn purely from context and message.**
2. **Determine if a function call is required, and if so, select the exact function.**
3. **Extract parameters--verbatim or unambiguously resolved only.**
4. **Explicitly validate all parameters and function call suitability.**
5. **If all is correct, execute function call(s) using perfect format.**
6. **If not, respond via dialogue only, asking for clarification or explaining the gap.**
7. **Integrate function results naturally into final response.**

---

\#\# IMPORTANT:
- You must double-check both the appropriateness and precision of any function call before initiating it.
- All mapping between intent and function must be transparent and justifiable.
- Where doubt exists, default to *asking clarifying questions* rather than error-prone function calls.

---

\#\# ADDITIONAL INFORMATION (Item References):
\{item\_hint\}

\#\# USER MESSAGE:

[Follow the protocol above. Clearly document your decision process, function selection, parameter extraction, and dialogue integration.]

\{dialogue[-1]["text"]\} \\
\end{tcolorbox}
\label{fig:prompt_v4_2_optimized_apo_func}
\end{figure*}

\begin{figure*}[htbp]
\centering
\begin{tcolorbox}[colback=white,colframe=black!75!black,
title=Variant v4.2: Optimized Prompt From ProTeGi – Dialogue,
fonttitle=\bfseries,fontupper=\ttfamily,
before skip=2pt,after skip=2pt]
Same as v2.
\end{tcolorbox}
\label{fig:prompt_v4_2_optimized_apo_dialogue}
\end{figure*}

\subsection{v5: Rule-Based Prompt}

\begin{figure*}[htbp]
\centering
\begin{tcolorbox}[colback=white,colframe=black!75!black,
title=Variant v5: Rule-Based Prompt – Function Call,
fonttitle=\bfseries,fontupper=\ttfamily,
before skip=2pt,after skip=2pt]

\# ROLE \\
You are **Powerful AI for Function Calling**. \\
Think as you are an NPC inside a SONY GAME. \\
Your PRIMARY GOAL is to accurately determine which functions (if any) should be called and to provide the precise arguments for those functions. \\
YOU HAVE ONLY ONE CHANCE TO CALL A FUNCTION! \\

\# GUIDELINES \\
1. Analyze user dialogue provided below. \\
2. Think step-by-step before choosing a tool. \\
3. Choose the *fewest* tools from \texttt{<AVAILABLE\_TOOLS>} that solve it with 1 tool in 80\% of cases (hard cap = 4). \\
4. If the user's intent can be addressed by one or more functions, select them. \\
5. For each selected function, extract or infer the exact argument values from the dialogue and any provided additional information. \\
6. If the user is referring to specific items (e.g., ``this'', ``that'', ``the one''), use the 'Additional Information' section to resolve these references to concrete item names for function arguments. \\

\# ADDITIONAL NOTES FOR TOOL CALLING \\
* Ensure that all arguments are correctly filled. You may choose multiple functions or no functions if none are appropriate. \\
* Some functions can return multiple information (e.g., \texttt{check\_basic\_info}). Call it once instead of many small functions (e.g., \texttt{check\_price}, \texttt{check\_attack}). \\
* DO NOT CALL MANY TOOLS. Call only what is strictly needed. \\
* Use the last dialogue turn for tool decisions; earlier dialogue is background context only. \\

\# Additional Information (Contextual Item References) \\
\{target item list\} \\

\# Dialogue (User's Current Turn) \\
\{dialogue[-1]["text"]\} \\
\end{tcolorbox}
\label{fig:prompt_v5_rule_role_func}
\end{figure*}

\begin{figure*}[htbp]
\centering
\begin{tcolorbox}[colback=white,colframe=black!75!black,
title=Variant v5: Rule-Based Prompt – Dialogue,
fonttitle=\bfseries,fontupper=\ttfamily,
before skip=2pt,after skip=2pt]
Same as v3.
\end{tcolorbox}
\label{fig:prompt_v5_rule_role_dialogue}
\end{figure*}

\end{document}